\documentclass[manuscript,screen,nonacm]{acmart}
\AtBeginDocument{%
  \providecommand\BibTeX{{%
    \normalfont B\kern-0.5em{\scshape i\kern-0.25em b}\kern-0.8em\TeX}}}

\acmConference[ACM Trans. Knowl. Discov.]{ACM Trans. Knowl. Discov.}{2021}{}
\acmBooktitle{ACM Trans. Knowl. Discov.}
\acmPrice{15.00}
\acmISBN{978-1-4503-XXXX-X/18/06}

\begin{document}

\title{Knowledge Enhanced Pretrained Language Models: A Compreshensive Survey}


\author{Xiaokai Wei}
\affiliation{%
  \institution{AWS AI}
  \city{New York}
  \country{USA}
}
\email{xiaokaiw@amazon.com}

\author{Shen Wang}
\affiliation{%
  \institution{AWS AI}
  \city{New York}
  \country{USA}
}
\email{shenwa@amazon.com}

\author{Dejiao Zhang}
\affiliation{%
  \institution{AWS AI}
  \city{New York}
  \country{USA}
}
\email{@dejiaoz@amazon.com}

\author{Parminder Bhatia}
\affiliation{%
  \institution{AWS AI}
  \city{New York}
  \country{USA}
}
\email{parmib@amazon.com}

\author{Andrew Arnold}
\affiliation{%
  \institution{AWS AI}
  \city{New York}
  \country{USA}
}
\email{anarnld@amazon.com}


\begin{abstract}
  Pretrained Language Models (PLM) have established a new paradigm through learning informative contextualized representations on large-scale text corpus. This new paradigm has revolutionized the entire field of natural language processing, and set the new state-of-the-art performance for a wide variety of NLP tasks. However, though PLMs could store certain knowledge/facts from training corpus, their knowledge awareness is still far from satisfactory. To address this issue, integrating knowledge into PLMs have recently become a very active research area and a variety of approaches have been developed. In this paper, we provide a comprehensive survey of the literature on this emerging and fast-growing field - Knowledge Enhanced Pretrained Language Models (KE-PLMs). We introduce three taxonomies to categorize existing work. Besides, we also survey the various NLU and NLG applications on which KE-PLM has demonstrated superior performance over vanilla PLMs. Finally, we discuss challenges that face KE-PLMs and also promising directions for future research.
\end{abstract}


%%\begin{CCSXML}
%%<ccs2012>
%% <concept>
%%  <concept_id>10010520.10010553.10010562</concept_id>
%%  <concept_desc>Computer systems organization~Embedded systems</concept_desc>
%%  <concept_significance>500</concept_significance>
%% </concept>
%% <concept>
%%  <concept_id>10010520.10010575.10010755</concept_id>
%%  <concept_desc>Computer systems organization~Redundancy</concept_desc>
%%  <concept_significance>300</concept_significance>
%% </concept>
%% <concept>
%%  <concept_id>10010520.10010553.10010554</concept_id>
%%  <concept_desc>Computer systems organization~Robotics</concept_desc>
%%  <concept_significance>100</concept_significance>
%% </concept>
%% <concept>
%%  <concept_id>10003033.10003083.10003095</concept_id>
%%  <concept_desc>Networks~Network reliability</concept_desc>
%%  <concept_significance>100</concept_significance>
%% </concept>
%%</ccs2012>
%%\end{CCSXML}

%\ccsdesc[500]{Computing methodologies~Natural language processing}
%\keywords{pretrained language model, knowledge graph, natural language processing}

\maketitle

\section{Introduction}

In recent years, large-scale pretrained language models (PLM), which are pretrained on huge text corpus typically with unsupervised objectives, have revolutionized the field of NLP. Pretrained models such as BERT \cite{DevlinJ2019}, RoBERTa \cite{LiuY2019}, GPT2/3 \cite{RadfordA2019} \cite{BrownT2020}  and T5 \cite{RaffelC2019} have gained huge success and greatly boosted state-of-the-art performance on various NLP applications \cite{QiuX2020}. The wide success of pretraining in NLP also inspires the adoption of self-supervised pretraining in other fields, such as graph representation learning \cite{HuW2020} \cite{HuZ2020} and recommender system \cite{SunF2019}\cite{XieX2020}.

%these models are still struggling on generation tasks that require reasoning over commonsense knowledge that is not explicitly stated in the context. For example, Figure 1 illustrates an example in the story ending generation task, where external commonsense knowledge in the form of relational paths can guide the generation of the key concepts “substance” and

% We refer interested readers to  for a comprehensive survey on pretrained models.
Training on large textual data also enables these PLMs memorize certain facts and knowledge contained in the training corpus. As demonstrated in recent work, these pretrained language models could possess decent amount of lexical knowledge \cite{LiuN2019} \cite{VulicI2020} as well as factual knowledge \cite{PetroniF2019} \cite{RobertsA2020} \cite{WangC2020}. However, further study reveals that PLMs also have the following limitations in terms of knowledge awareness:
\begin{itemize}
    \item For NLU, recent study have found PLMs tend to rely on superficial signals/statistical cues \cite{PetroniF2020} \cite{McCoyT2019} \cite{NivenT2019}, and can be easily fooled with negated (e.g., ``Birds can [MASK]'' v.s. ``Birds cannot [MASK]'') and misprimed probes \cite{KassnerN2020}. Besides, It has been found that PLMs often fail in reasoning tasks \cite{TalmorA2020}.
    \item For NLG, although PLMs are able to generate grammatically correct sentence, the generated text might not be logical or sensible \cite{LinB2020} \cite{LiuY2020}. For example, as noted in \cite{LinB2020}, given a set of concepts \{dog, frisbee, catch, throw\}, GPT2 generates ``\textit{A dog throws a frisbee at a football player}'' and T5 generates ``\textit{dog catches a frisbee and throws it to a dog}'', neither of which aligns with human's commonsense.
    % \item Knowledge might need to dynamically evolve and new knowledge replaces outdated knowledge 
\end{itemize}

These observations motivate work on designing more knowledge-aware pre-trained models. Recently, an ever-growing body of work aims at explicitly incorporating knowledge into PLMs \cite{YamadaI2020} \cite{ZhangZ2019}\cite{PetersM2019}\cite{VergaP2020}\cite{WangR2020}\cite{LiuW2020}\cite{JiH2020}. They exploit knowledge from various sources such as encyclopedia knowledge, commonsense knowledge and linguistic knowledge with different injection strategies. Such knowledge integration mechanism have successfully enhance existing PLMs' knowledge awareness, and lead to improved performance on a variety of tasks, including but not limited to entity typing \cite{YamadaI2020}, question answering \cite{YangA2019}\cite{LinY2019}, story generation \cite{GuanJ2020} and knowledge graph completion \cite{YaoL2019}.

%An effective way to making the PLMs more knowledge aware is to inject knowledge into these models. 

% Nevertheless, open questions remain about what these models have learned and improvements can be made along several directions. One such direction is, when downstream task performance depends on structured relational knowledge – the kind modeled by knowledge graphs (KGs) – 

%  some efforts have been mawhich can pose a problem for effective graph mining.

%show pre-trained LMs have been partially equipped with such knowledge. Interestingly, it has also recently been observed that these models can internalize a sort of implicit “knowledge base” after pre-training 

%Recently, a new approach has emerged to address the above-mentioned problem . An attention mechanism aids a model by allowing it to "focus on the most relevant parts of the input to make decisions" [Velickovic et al. 2018]. Attention was first introduced in the deep learning community to help models attend to important parts of the data [Bahdanau et al. 2015; Mnih et al. 2014]. 

%considering the millions of documents and facts in Wikipedia1 and other textual resources, it unlikely that a language model with a fixed number of parameters is able to reliably store and retrieve factual knowledge with sufficient precision 

\begin{figure*}[t]
\includegraphics[width=1.23\textwidth, height=0.59\textheight]{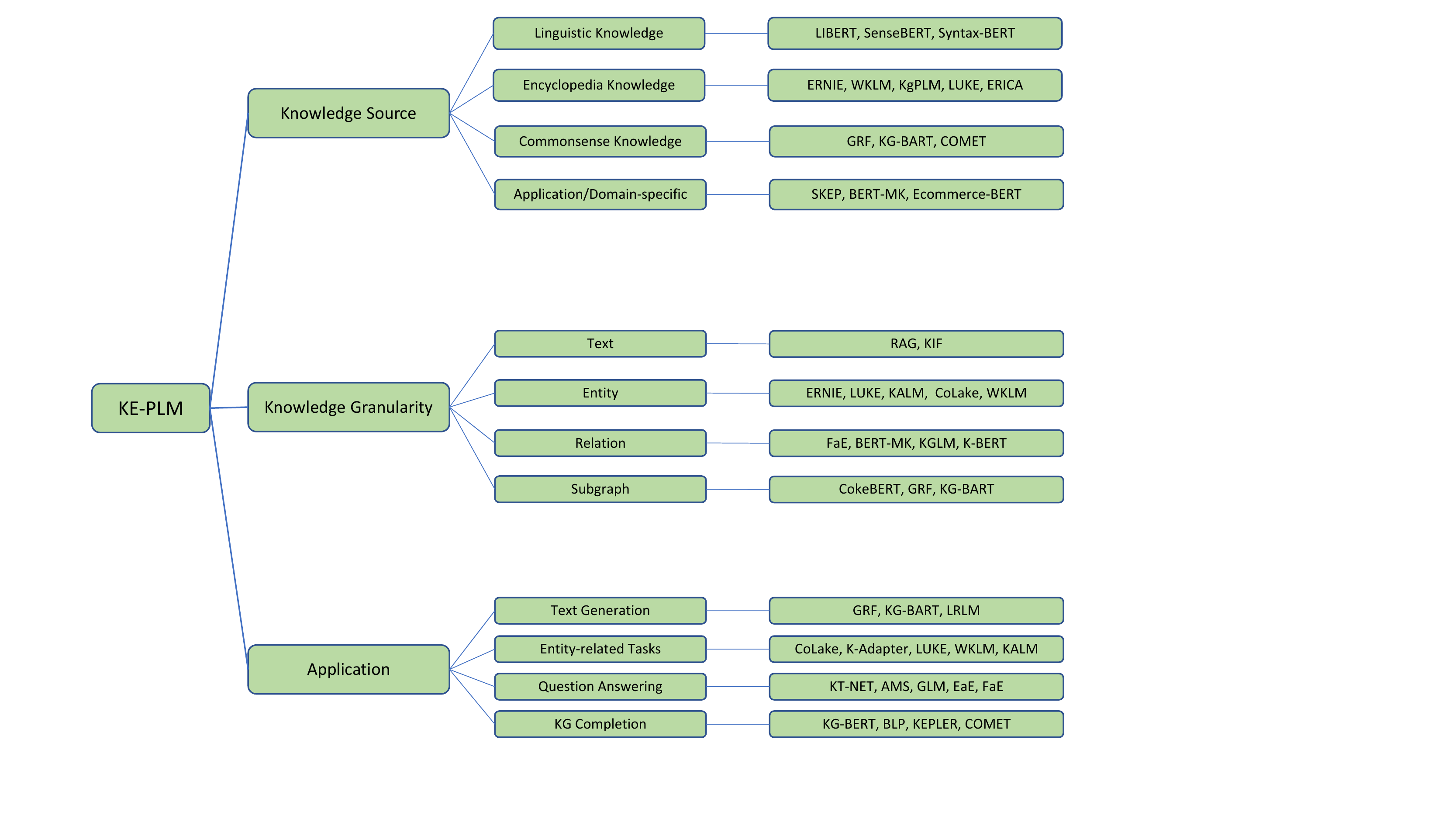}
\centering
\caption{Taxonomy of Knowledge Enhanced Pretrained Langauge Models (KE-PLMs)}
\label{fig:category}
\end{figure*}

In this paper, we aim to provide a comprehensive survey on this emerging field of Knowledge Enhanced Pretrained Language Models (KE-PLMs). Existing work on KE-PLMs have developed a diverse set of techniques for knowledge integration on different knowledge sources. To provide insights on these models and facilitate future research, we build three taxonomies to categorize the existing KE-PLMs. Figure \ref{fig:category} illustrates our proposed taxonomies on Knowledge Enhanced Pretrained Language Models (KE-PLMs). 

In existing KE-PLMs, there are different types of knowledge sources (e.g., linguistic, commonsense, encyclopedia, application-specific) have been explored to enhance the capability of PLMs in different aspects. The first taxonomy helps us to understand what knowledge sources have been considered for constructing KE-PLMs. In the second taxonomy, we recognize that a knowledge source can be exploited to different extents, and categorize the existing work based on the knowledge granularity: text-chunk based, entity-based, relation triple-based and subgraph-based. Finally, we introduce a third taxonomy that groups the methods by their application areas. This taxonomy presents a range of applications that existing KE-PLMs have targeted to improve with the help of knowledge integration. By recognizing what application areas have been well addressed/under addressed by KE-PLMs, we believe this could shed light on future research opportunities on applying KE-PLMs to under-addressed areas.

% Most previous works (as shown in Table 1) augment the standard language modeling objective with knowledge-driven objectives and update the entire model parameters. Although these methods obtain better performance on downstream tasks, they struggle at supporting the development of versatile models with multiple kinds of knowledge injected (Kirkpatrick et al., 2017). When new kinds of knowledge are injected, model parameters need to be retrained so that previously injected knowledge would fade away

%We organized Sections 3-5 such that the methods in existing work are grouped by the main type of embedding they calculate (e.g., node embedding, edge embedding, graph embedding, or hybrid embedding); these methods are then further divided by the type of graphs they support. In Sections 6 and 7, we switch to a different perspective and use the remaining taxonomies (Fig. 2b and Fig. 2c) to guide the discussion. We then discuss challenges as well as interesting opportunities for future work in Section 8. Finally, we conclude the survey

% In this survey, we present a systematic survey on these knowledge enhanced language models. We summarize the applications and tasks on which existing work on KE-PLMs aims to improve. 

{\bf Related work} Two contemporaneous reviews \cite{SafaviT2021} and \cite{HernandezP2021} also investigate incorporating relational knowledge graph into pretrained language models. We cover a larger scope by discussing more types of knowledge (e.g., domain specific knowledge) and present in-depth discussion on the applications and potential future directions.

The rest of this survey is organized as follows. In section \ref{knowledge_source}, we present our taxonomy based on the knowledge source of KE-PLMs, and discuss representative approaches in each category. In section \ref{knowledge_granularity}, we categorize the different knowledge granularities that existing KE-PLMs exploit, and summarize the common techniques employed for incorporating such knowledge. In section \ref{application}, we present the applications that benefit from the development of KE-PLMs and introduce corresponding datasets. In section \ref{futuredirection}, we discuss the challenges facing the design of highly effective KE-PLMs and the opportunities for future research in this area. Lastly, we conclude our survey in \ref{conclude}.

%Let $\mathcal{G}=(\mathcal{E},\mathcal{R})$ be a KG with multiple relations, where $\mathcal{E}$ and $\mathcal{R}$ represents the set of entities (nodes) and relations (edges), respectively. A triple $(e_h,r,e_t) \in \mathcal{E} \times \mathcal{R} \times \mathcal{E}$ is represented as an edge $r$ between head entity $e_h$ and tail entity $e_t$ in $\mathcal{G}$.
% A knowledge graph is a multi-relational graph denoted by $\mathcal{G}=(\mathcal{E},\mathcal{R})$, where $\mathcal{E}$ and $\mathcal{R}$ represents the set of entities (nodes) and relations (edges), respectively. A triple $(e_h,r,e_t) \in \mathcal{E} \times \mathcal{
% R} \times \mathcal{E}$ is represented as an edge $r$ between head entity $e_h$ and tail entity $e_t$ in $\mathcal{G}$.

\section{Knowledge Source} \label{knowledge_source}

Existing works have explored the integration of knowledge from diverse sources: linguistic knowledge, encyclopedia knowledge, and commensense knowledge and domain-specific knowledge. In this section, we categorize existing work by their knowledge source. For each category of knowledge source, we also introduce several representative methods and the corresponding knowledge they exploit.

\subsection{Linguistic Knowledge} \label{linguistic_knowledge}

\textbf{Lexcial relation} Lexically Informed BERT (LIBERT) \cite{LauscherA2020} incorporate lexical relation knowledge by predicting whether two words in a sentence exhibit semantic similarity (i.e., whether they are synonyms or hypernym-hyponym pairs).

\textbf{Word sense} SenseBERT \cite{LevineY2020} exploits word-sense knowledge from WordNet as weak supervision by including additional training task on predicting word-supersense (e.g., \textit{noun.food} and \textit{noun.state}) based on the masked word’s context.

\textbf{Syntax tree} Syntax-BERT\cite{BaiJ2021} employs syntax parsers to extract both dependency \cite{ChenD2014} and constituency parsing \cite{ZhuM2013} trees. Syntax-related masks are designed to incorporate information from the syntax trees. K-Adapter\cite{WangR2020} also includes a linguistic adapter that incorporates dependency parsing information, by predicting the head index for each token in a sentence.

\textbf{Part-of-Speech tag} SentiLARE \cite{KeP2020} and LIMIT-BERT \cite{ZhouJ2020} consider POS tag as additional knowledge. For example, LIMIT-BERT incorporates multiple types of linguistic knowledge simultaneously in a multi-task manner. In addition to POS tags and syntactical parsing tree, LIMIT-BERT also explores span and semantic role labeling (SRL) information.
 
With such additional linguistic knowledge incorporated, these approaches are able to demonstrate superior performance on general NLU benchmark dataset such as GLUE \cite{WangA2018} and SuperGLUE\cite{WangA2019}, or specific applications such as sentiment analysis.

\subsection{Encyclopedia Knowledge}

Encyclopedia KG such as Freebase \cite{BollackerK2008}, NELL \cite{CarlsonA10} and Wikidata \cite{VrandecicD2014} contain facts/world knowledge in the following form:
\begin{equation}
    (head\_entity, relation, tail\_entity) \notag
\end{equation}
For example, \textit{('Joe Biden', 'PresidentOf', 'USA')}. These encyclopedia KG are able to provide abundant knowledge for PLMs to integrate. A majority of exiting work on KE-PLMs\cite{ZhangZ2019}\cite{HeB2020b}\cite{YuD2020} \cite{WangR2020}\cite{VergaP2020}\cite{SunT2020}\cite{QinY2020} uses Wikidata\footnote{https://www.wikidata.org/} as knowledge source. Typically, entities in Wikidata are linked with entity mentions in the text of Wikipedia. Then entity-aware training could be performed on such linked data to learn the parameters for KE-PLMs.

\subsection{Commonsense Knowledge}

To empower PLMs with more commonsense reasoning capability, existing models typically resort to the following two commonsense knowledge graphs: ConceptNet \cite{SpeerR2017} and ATOMIC \cite{SapM2019}.
\begin{itemize}
\item ConceptNet\footnote{https://conceptnet.io/} \cite{SpeerR2017} is a multilingual knowledge graph consisting of triples in $34$ relations, such as \textit{CapableOf}, \textit{Causes} and \textit{HasProperty}. For a knowledge triple $(concept1, r, concept2)$, it represents that head $concept1$ has the relation $r$ with tail $concept2$. For example, the triple \textit{(cooking, requires, food)} means that “the prerequisite of cooking is food”. 
\item ATOMIC \cite{SapM2019} contains inferential knowledge in the form \textit{if-then} triples. It covers a variety of social commonsense knowledge around specific event prompts (e.g., “X goes to the store”). Specifically, ATOMIC distills its commonsense in nine dimensions, covering the event’s causes and effects on the agent.
\end{itemize}
Incorporating knowledge from ConceptNet and ATOMIC helps PLMs gain stronger capability on commonsense reasoning \cite{LinY2019}\cite{ShenT2020}\cite{YeZ2019}\cite{LvS2020}  \cite{GuanJ2020}\cite{JiH2020}\cite{YuC2020}\cite{LiuY2020}\cite{LiY2020}. We will discuss in more detail about how the knowledge is incorporated in section \ref{knowledge_granularity} and how they benefit multiple commonsense-related downstream tasks such commonsenseQA and text generation in section \ref{application}.

\subsection{Application/Domain Specific Knowledge}
In this subsection, we introduce KE-PLMs which utilizes domain-specific knowledge for their particular vertical applications.

\textbf{Sentiment knowledge} In addition to the POS information mentioned in section \ref{linguistic_knowledge}, SentiLARE\cite{KeP2020} also utilizes sentiment word polarity from SentiWordNet\cite{BaccianellaS2010}. SKEP\cite{TianH2020} incorporates sentiment knowledge from self-supervised training, including sentiment word detection, word polarity and aspect-sentiment pair.

\textbf{Medical knowledge}
Medical domain knowledge \cite{HeB2020a} integrate biomedical ontology from Unified Medical Language System (UMLS) \cite{BodenreiderO2004} to facilitate tasks in medical domain. K-BERT \cite{LiuW2020} also exploits knowledge from a medical concept KG for higher quality NER in medical domain.

\textbf{E-commerce Product Graph} E(commerce)-BERT\cite{ZhangD2020} utilizes product association graph (i.e., whether two products are substitutable and complementary) \cite{McAuleyJ2015} constructed from consumer shopping statistics. It introduces additional tasks for reconstructing a product given its neighbor products in association graph.

Though most existing approaches exploit only one knowledge source, it is worth noting that certain methods attempt to incorporate from more than one knowledge source. For example, K-Adapter \cite{WangR2020} incorporate knowledge from multiple sources by learning a different adapter for each knowledge source. It exploits both dependency relation as linguistic knowledge and relation/fact knowledge from Wikidata.

\section{Knowledge Granularity} \label{knowledge_granularity}

A majority of approaches resort to knowledge graphs (enclopedia, commonsense or domain-specific) as source of knowledge. In this section, we group these models by the granularity of knowledge they incorporate from KG: text-based knowledge, entity knowledge, relation triples and KG subgraphs.

\subsection{Text Chunk-based Knowledge}

RAG \cite{LewisP2020} builds an index in which each entry is a text chunk from Wikipedia. It first retrieve top-k documents/chunks from the memory using kNN-based retrieval, and the BART model \cite{LewisM2020} is employed to generate the output, conditioned on these retrieved documents. Similarly, KIF \cite{FanA2020} uses KNN-based retrieval on an external non-parametric memory storing wikipedia sentences and dialogue utterances to improve generative dialog modeling.

\subsection{Entity Knowledge} \label{entity_knowledge}

Entity-level information can be highly useful for a variety of entity-related tasks such as NER, entity typing, relation classification and machine reading comprehension. Hence, many existing KE-PLM models target this type of simple yet powerful knowledge.

A popular approach to making PLMs more entity aware is to introduce the entity-aware objectives while pretraining. Such strategy is adopted by multiple existing approaches: ERNIE (BAIDU)\cite{SunY2020} ERNIE(THU)\cite{ZhangZ2019} CokeBERT\cite{SuY2020} KgPLM\cite{HeB2020b}, LUKE\cite{YamadaI2020}, GLM\cite{ShenT2020}, KALM\cite{RossetC2020}, CoLAKE \cite{SunT2020}, JAKET\cite{YuD2020} and AMS\cite{YeZ2019}. A typical choice of entity-related objective is an entity linking loss which predicts the entity mention in text to entity in KG with a cross entropy loss or max-margin loss on the prediction \cite{ZhangZ2019}\cite{YamadaI2020}\cite{YuD2020}\cite{RossetC2020}. There is also method that employs replacement detection loss in which the goal is the predict whether an entity mention has been replaced with a distractor \cite{XiongW2020}. Also, certain methods adopts more than one type of losses. For example, KgPLM\cite{HeB2020b} show through ablation study that employing both a cross-entropy entity prediction loss and an entity replacement detection loss (which were referred to as generative and discriminative loss respectively) lead to better performance than only using one loss. We summarize the different entity-aware loss functions in Table \ref{table:entity_loss}.

\begin{table}
\begin{center}
    \caption{Summarization of entity-related objectives. $s_i$ is the similarity score between mention $m_i$ and entity $e_i$ (e.g., the inner product between them) given the contextulized embedding $\mathbb{C}$. $s^{-}_i$ is the similarity on negative(i.e. distractor) mention/entity pairs. $r_i$ represents whether the entity has been replaced with any negative sample.}
    \begin{tabular}{| l | l |  l | }
    \hline
    Loss Type & Typical Form & Reprensentative Methods \\ \hline
    \shortstack{ Cross entropy  \\ entity linking (EL) loss } & $\sum_i \mathbb{I}(m_i=e_i)\log(e_i|\mathbb{C})$  &  \shortstack{ ERNIE(THU)\cite{ZhangZ2019}, JAKET\cite{YuD2020} \\ EaE\cite{FevryT2020}, LUKE\cite{YamadaI2020}, CokeBERT\cite{SuY2020}}  \\ \hline
    Max margin EL loss  & $\sum_i max(\lambda- s_i + s^{-}_i, 0)$  &   GLM\cite{ShenT2020}, KALM\cite{RossetC2020}  \\ \hline
    Entity replacement detection & \shortstack{  $\sum_j \mathbb{I}(r_i=1)\cdot \log p(r_i=1|\mathbb{C})$ + \\ $\mathbb{I}(r_i=0)\cdot \log p(r_i=0|\mathbb{C}) $}  &    WKLM\cite{XiongW2020}, GLM\cite{ShenT2020}  \\ \hline
    Squred Loss & $ \sum_i || \mathbf{W}e_i - m_i||^2$ & E-BERT \cite{PoernerN2020} \\ \hline
    \end{tabular}
    \label{table:entity_loss}
\end{center}
\end{table}

%\begin{mydef}
%{\bf SideSim} Given a side information network, we define the following similarity measure from the side information w.r.t meta-path $m \in M$ as follows:
%\end{mydef}.

%In additional to instroducing entity-related loss function, certain methods also explore 

%E-BERT \cite{PoernerN2020} propose two variants for combining entity information and text token: E-BERT-concat concatenates the entity ID and text token, while E-BERT-replace replaces the text token with entity vector. KT-NET \cite{YangA2019} generates knowlege-enhanced embedding by applying attention mechanism on extracted entities, and they should . % Rather than using an off-the-shelf tool for entity extraction and linking, EaE has a built-in detection and linking layer by including corresponding objectives in pre-training. 

Besides named entities, a few existing approaches find it helpful to incorporate phrase information into pretraining, such as ERNIE (BAIDU) \cite{SunY2020} and E(commerce)-BERT\cite{ZhangD2020}.

\subsection{Relation Knowledge} \label{relation_knowledge}

KE-PLMs by incorporating entity information alone have already demonstrated impressive performance gains over vanilla PLMs. Recently, an increasing amount of work have considered integrating knowledge beyond entities, such as the relation triples in KG to make the models even more powerful. 

For example, FaE (Fact-as-Experts) \cite{VergaP2020} further extends EaE \cite{FevryT2020} by building fact (i.e., relation triples in KG) memory in addition to the entity memory. The set of tail entities retrieved from fact memory are aggregated with attention mechanism, and then fused with token embeddings. BERT-MK \cite{HeB2020a} and adopts similar technique as ERNIE-THU\cite{ZhangZ2019} for knowledge fusion. ERICA\cite{QinY2020} applies contrastive loss on entity and relation, which pulls neighbor entity/relation close and pushes non-neighbors far apart in the embedding space. K-Adapter\cite{WangR2020} introduces a factual adapter which incorporates relation information by performing relation classification based on the entity context. K-BERT\cite{LiuW2020} injects knowledge by augmenting sentence with the triplets from KG to transform it into a knowledge-rich sentence tree. To prevent changing the original semantic of the sentence, K-BERT adopts a visible matrix to control the visibility of each injected token.

To better predict the plausibility of a triple, KG-BERT \cite{YaoL2019} fine-tune a BERT model on a training format as in (\ref{kg_bert}) derived from KG triplets. Similarly, COMET \cite{BosselutA2020} and \cite{GuanJ2020} verbalize the concepts and relations from commonsense KG into text tokens for training a knowledge enhanced PLM. KEPLER \cite{WangX2019} and BLP (BERT for Link Prediction) \cite{DazaD2020} further include a TransE based objective \cite{BordesA2013} for the PLM to incorporate information from the triples in KG. 
\begin{equation}
    [CLS]\text{HeadEntityTokens}[SEP]\text{ReltionTokens}[SEP]\text{TailEntityTokens}[SEP]
    \label{kg_bert}
\end{equation}

LRLM \cite{HayashiH2019} and KGLM \cite{LoganR2019} are two language models that utlize relations in KG when generating the next token in an autoregressive manner. They introduce a latent variable to determine whether the next token should be a text-based token or entity/relation induced token. 

\subsection{Subgraph Knowledge}

As knowledge graphs provide richer knowledge than simply entity/relation triplets, recently, an increasing number of approaches start to explore integration of more sophisticated knowledge, such as subgraphs in KG. The approaches in this category typically have two stages: subgraph construction and subgraph representation learning. In the subgraph construction stage, they extract the 1-hop or K-hop relations. As the subgraph can be large (particularly when $K\ge 2$) and many entities/relations involved might not be highly relevant to the text context, usually certain pruning/filtering are conducted and/or attention mechanism are applied on the nodes/relations. In the second stage, for representation learning on extracted subgraphs, most existing approaches adopt certain variants of Graph Neural Networks with attention mechanism.

% To incorporate KG subgraphs, a key challenge is to effectively represent the subgraphs. 

\subsubsection{Encylopedia KG}

CokeBERT \cite{SuY2020} designs Semantic-driven GNN (S-GNN) for representing knowledge subgraphs. S-GNN employs attention mechanism to direct the model to better focus on the most relevant entities/relations. Subsequently, conditioned on pre-trained entity embedding and subgraph embedding, the entity mention embedding is computed and used in a way similar to ERNIE (THU) \cite{ZhangZ2019}.

CoLake \cite{SunT2020} also uses GNN to aggregate information from the extracted knowledge subgraphs in both pretraining and inference. Instead of constructing GNN in an explicit way, CoLake convert the subgraph into token sequence and append it to input sequence for PLM, by recognizing the fact that self-attention mechanism in Transformer \cite{VaswaniA2017} is similar to Graph Attention Networks \cite{VelickovicP2018} in spirit.

\subsubsection{Commonsense KG}

Knowledge subgraphs are also very useful for commonsense based applications, such as CommonsenseQA and text/story generation.

Kag-Net \cite{LinY2019} employs path-finding based algorithm for subgraph construction. It first applies GCN on the subgraph to generate node embeddings, then the output of GCNs is passed to LSTM-based \cite{HochreiterS1997} path encoder. Finally, the path embeddings on the subgraphs are combined through attention-based pooling to obtain the knowledge representation.

GRF \cite{JiH2020} also adopts a GNN-based module for concept subgraph encoding, and it designs a dynamic reasoning module to propogate information on this graph at each decoding step.

KG-BART \cite{LiuY2020} first constructs a knowledge subgraph from commonsense KG, and use Glove embedding \cite{PenningtonJ2014} based word similarity to prune irrelevant concepts. Then graph attention is employed to learn concept embeddings, which are integrated with token embedding using concept-to-subword and subword-to-concept fusion. Such scheme enables KG-BART to generate sentences with more logical and natural even with unseen concepts. \cite{LvS2020} also uses GNN to generate embedding for concepts graphs.

In Table \ref{table:model_compare}, we summarize the existing approaches with their characteristics.

\begin{table}
\begin{center}
    \caption{Comparisons of different knowledge enhanced pretrained langauge models}
    \begin{tabular}{| l | l | l | l |  }
    \hline
    Method & \shortstack{Knowledge \\ Aware \\ Pretraining} & \shortstack{Knowledge \\Aware \\Auxiliary Loss} & Knowledge Source \\ \hline
    ERICA \cite{QinY2020} & Yes  & entity/relation discrimination  &  Wikipedia, Wikidata   \\ \hline
    ERNIE (THU) \cite{ZhangZ2019} & Yes  & entity prediction & Wikipedia/Wikidata  \\ \hline
    ERNIE 2.0 (Baidu) \cite{SunY2020} & Yes   & masked entity/phrase  &  N/A  \\ \hline
    E-BERT \cite{PoernerN2020} &  Yes  & entity/wordpiece alignment   &  Wikipedia2Vec  \\ \hline
    E(commerce)-BERT \cite{ZhangD2020} &  Yes  & neighbor Product Reconstruction   &  product graph/AutoPhrase\cite{ShangJ2018}  \\ \hline
    EaE \cite{FevryT2020} &  Yes  & mention detection/linking  & Wikipedia \\  \hline
    CokeBERT \cite{SuY2020} &  Yes  &  entity prediction  & Wikipedia/Wikidata \\ \hline
    COMET \cite{BosselutA2020}  & No   & autoregressive   &  ATOMIC, ConceptNet  \\ \hline
    K-Adapter \cite{WangR2020} &  No  & dependency relation   &  Wikipedia, Wikidata, Stanford Parser   \\ \hline
    KnowBERT \cite{PetersM2019} &  Yes  & entity linking   &  WordNet, Wikipedia  \\ \hline
    K-BERT \cite{LiuW2020} &  No  & finetuning   & \shortstack{ WikiZh, WebtextZh, CN-DBpedia \\ HowNet, MedicalKG } \\ \hline
    KEPLER \cite{WangX2019} &  Yes  & TransE scoring   &   Wikipedia/Wikidata \\ \hline
    KG-BERT \cite{YaoL2019} &  Yes  &  relation cross-entropy  &  ConceptNet   \\ \hline 
    KG-BART \cite{LiuY2020} &  Yes  &  masked concept  &  ConceptNet   \\ \hline
    KgPLM \cite{HeB2020b} &  Yes  & generative/discriminative masked entity   &  Wikipedia/Wikidata  \\ \hline
    FaE \cite{VergaP2020} &  Yes  & masked entity et.al   &  Wikipedia/Wikidata \\ \hline
    JAKET \cite{YuD2020} &  Yes  & entity category/relation type/masked entity   &  Wikipedia/Wikidata \\ \hline
    LUKE \cite{YamadaI2020} &  Yes  &  entity prediction   &  Wikipedia  \\ \hline
    WKLM \cite{XiongW2020} & Yes & entity replacement detection & Wikipedia/Wikidata \\ \hline
    CoLAKE \cite{SunT2020} &  Yes  &  masked entity prediction   &  Wikipedia/Wikidata   \\ \hline
    KT-NET \cite{YangA2019} &  No  &  finetuning  &  N/A \\ \hline
    LIBERT \cite{LauscherA2020b} &  Yes  & lexical relation prediction   &  WordNet  \\ \hline
    SenseBERT \cite{LevineY2020} & Yes   & supersense prediction  &  WordNet   \\ \hline
    Syntax-BERT \cite{BaiJ2021} &  No  & masks induced by syntax tree parsing   & syntax tree  \\ \hline
    SentiLARE \cite{KeP2020} &  Yes  &  POS/ word level polarity/sentiment polarity   &  SentiWordNet   \\ \hline
    \cite{LiY2020}  &  No  &  finetuning  &  ConceptNet   \\ \hline
    COCOLM \cite{YuC2020} &  Yes  & discourse relation/co-occurrence relation   &  ASER   \\ \hline
    \cite{GuanJ2020} &  Yes  &  autoregressive  &  ConceptNet/ATOMIC  \\ \hline
    AMS \cite{YeZ2019} &  Yes  &  distractor-based loss  &  ConceptNet   \\ \hline
    GLM \cite{ShenT2020} &  No  & distractor-based entity linking loss  &  ConceptNet \\ \hline
    GRF \cite{JiH2020} & No   &  finetuning  & ConceptNet    \\ \hline
    KagNet \cite{LinY2019} &  No  & finetuning   &  ConceptNet   \\ \hline
    LIMIT-BERT \cite{ZhouJ2020} &  Yes  & semantics/syntax   & pretrained model  \\ \hline
    KGLM \cite{LoganR2019} &  Yes  &  autoregressive  &  WikiText-2/wikidata   \\ \hline
    LRLM \cite{HayashiH2019} &  Yes  & autoregressive   &  Wikidata/Freebase  \\ \hline
    \cite{OstendorffM2019} &  No  &  finetune  &  Wikidata \\ \hline
    SKEP \cite{TianH2020} &  Yes  & Word Polarity Prediction   &  auto-mined  \\ \hline
    \hline
    \end{tabular}
    \label{table:model_compare}
\end{center}
\end{table}

\section{Applications} \label{application}

Knowledge enhanced pretrained language models have benefited a variety of NLP applications, especially those knowledge-intensive ones. In the following, we discuss in more detail how KE-PLM improves the performance of various NLG and NLU tasks. To facilitate future research in the area, we also briefly introduce a few widely used benchmark datasets for evaluating the efficacy of these models.

\subsection{Text Generation}

The KE-PLMs designed for natural language generation can be roughly categorized into following two types:
\begin{itemize}
    \item To improve the logical soundness of generated text by incorporating commonsense knowledge graphs.
    \item To improve the factualness of generated text by incorporating encyclopedia knowledge 
\end{itemize}
In the following, we introduce the commonly used datasets for evaluating these two aspects.

ROCStories \cite{MostafazadehN2016} consists of $98,162$ coherent stories (each with five sentences) as the unlabeled training dataset. ROCStories is widely used for story understanding and generation tasks. By integrating subgraph information from Commonsense KG (e.g., ConceptNet and ATOMIC), knowledge enhanced PLMs \cite{GuanJ2020}\cite{JiH2020}\cite{YuC2020} are able to generate story endings that is more logical and aligns better with the commonsense of human.

CommonGen \cite{LinB2020} is a popular dataset for constrained text generation task. It combines crowdsourced and existing caption corpora to produce commonsense descriptions for over 35k concept sets. The goal of CommonGen is to assess the model's capability on commonsense reasoning when generating text. In this task, given a concept set with typically three to five concepts, the model is expected to generate coherent and sensible text description containing these concepts. KG-BART \cite{LiuY2020} and \cite{LiY2020}, via incorporating commonsense knowledge, show better capability on generating more natural and sensible text for the given concept set.

% \subsubsection{Factualness Evaluation}

\subsection{Entity-related Tasks}

Since many existing approaches incorporate entity-level knowledge, entity related tasks (e.g., entity typing and relation classification) become natural testbeds for evaluating the efficacy of these KE-PLMs. By injecting entity information into the model, they are able to be more entity-aware and outperform these vanilla PLMs (such as BERT and RoBERTa) on different benchmark datasets.

\subsubsection{Entity Typing}

The goal of entity typing is to classify entity mentions to predefined categories. Two popular benchmark datasets used for entity typing are Open Entity\cite{ChoiE2018} and FIGER\cite{LingX2015} (sentence-level entity typing dataset)). To perform entity typing, existing approaches \cite{SuY2020}\cite{WangX2019}\cite{ZhangZ2019}\cite{YamadaI2020}\cite{SunT2020}\cite{PoernerN2020}\cite{WangR2020} typically insert special tokens [E] and [/E] to surround the entity mention and employ the contextual representation of the [E] token to predict the category.

\subsubsection{Relation Classification}

Relation classification/relation typing aims at classifying the relationship between two entities mentioned in text. TACRED\cite{ZhangY2017} is widely used benchmark dataset for relation classification. It consists of over $10k$ sentences with a total of 42 relations. FewRel\cite{HanX2018} is a dataset constructed from Wikipedia text and Wikidata facts. It contains $100$ relations can be used for evaluating few-shot relation classification. 

To perform relation classification task, existing PLMs \cite{SuY2020}\cite{ZhangZ2019}\cite{WangX2019}\cite{ZhangZ2019}\cite{YamadaI2020}\cite{SunT2020}\cite{PoernerN2020}\cite{WangR2020} typically perform the following during fine-tuning: special tokens [HE], [/HE], [TE] and [/TE] are added to mark the beginning and end for head entity and tail entity. Then they usually concatenate the contextual embeddings for [HE] and [TE] to predict the relation category.

\subsection{Question Answering}
\subsubsection{Cloze-style Question Answering/Knowledge Probing}

LAMA (LAnguage Model Analysis) probe \cite{PetroniF2019} is a set of cloze-style questions with single-token answers. It is generated from the relation triplets in KG with templates which contain variables \textit{s} and \textit{o} for subject and object (e.g, ``\textit{s} was born in \textit{o}''). This dataset aims at measuring how much factual knowledge is stored in pretrained models. \cite{PoernerN2020} further constructs LAMA-UHN, a more difficult subset of LAMA, by filtering out easy-to-answer questions with overly informative entity names. 

CoLAKE\cite{SunT2020}, E-BERT\cite{PoernerN2020}, KgPLM\cite{HeB2020b}, EaE \cite{FevryT2020}, KALM\cite{RossetC2020}, KEPLER\cite{WangX2019} and K-Adapter\cite{WangR2020} adopt LAMA for knowledge probing. With the injected knowledge, they are able to generate more factual answers than vanilla PLMs.

% To better design effective knowledge fusing techniques for PLMs, it is helpful to probe/examine how well PLMs 

\subsubsection{Open Domain Question Answering}

PLMs equiped with Knowledge could lead to better performance on Open Domain Question Answering (ODQA), as the context and answer often involve entities. A few KE-PLMs evaluate their approaches on ODQA dataset to showcase their improved capability. For example, KT-NET\cite{YangA2019} and LUKE\cite{YamadaI2020} demonstrate their efficacy on SQuAD1.1\cite{RajpurkarP2016}. TriviaQA\cite{JoshiM2017} and SearchQA\cite{DunnM2017} were used by EaE\cite{FevryT2020} and K-ADAPTER\cite{WangR2020}, respectively.

% \cite{VergaP2020}, FreebaseQA, WebQuestionsSP

\subsubsection{Commonsense QA}

CommonsenseQA \cite{TalmorA2019} is a dataset question answering constructed from ConceptNet \cite{SpeerR2017} to test model's capability on understanding several commonsense types (e.g., causal, spatial, social). Each question is equipped with one correct answer and four distractors by human annotators. The distractors are typically also related to the question but less aligned with human commonsense. One example question is "what do all humans want to experience in their own home? \{ \textbf{feel comfortable}, work hard, fall in love, lay eggs, live forever\}" \footnote{https://www.tau-nlp.org/commonsenseqa}. Through integrating knowledge from commonsense KG, KagNet\cite{LinY2019}, GLM\cite{ShenT2020}, AMS\cite{YeZ2019} and \cite{LvS2020} are able to perform reasonably well on CommonsenseQA.  CosmosQA\cite{HuangL2019} is a dataset with multiple-choice questions for commonsense-based reading comprehension, and K-Adapter\cite{WangR2020} outperforms vanilla RoBERTa\cite{LiuY2019} on this dataset.

\subsection{Knowledge Graph Completion}

Knowledge graphs usually suffer from the problem of incompleteness and many relations between entities are missing in KG. This might negatively impact multiple downstream tasks, such as knowledge base QA \cite{HaoY2017}. Pretrained language models enhanced by knowledge graph can in turn help infer missing links and improve the completeness of knowledge graphs. Some popular benchmark datasets for evaluating link prediction on knowledge graphs are WN18RR \cite{DettmersT2018} (subset of WordNet), FB15K-237 \cite{DettmersT2018} (subset of FreeBase \cite{BollackerK2008}) and also Wikidata5M (recently introduced in \cite{WangX2019}).

KG-BERT \cite{YaoL2019}, BLP \cite{DazaD2020}, GLM \cite{ShenT2020} and KEPLER \cite{WangX2019} by including margin-based or cross-entropy loss derived from relation triplet, demonstrated competitive performance on the datasets mentioned above. Besides, it has been shown that COMET \cite{BosselutA2020} is able to generate new high-quality commonsense knowledge, and hence can be employed to further complete commonsense KGs such as ConceptNet \cite{SpeerR2017} and ATOMIC \cite{SapM2019}.

\subsection{Sentiment Analysis Tasks}

KE-PLMs also demonstrate their effectiveness on sentiment analysis tasks such as Amazon QA dataset \cite{MillerJ2020}, Stanford Sentiment Treebank (SST) \cite{SocherR2013}, Semantic Eval 2014 Task4 \cite{PontikiM2014}. With the help of additional knowledge such as word sentiment polarity, SentiLARE \cite{KeP2020}, SKEP \cite{TianH2020}, E(commerce)-BERT \cite{ZhangD2020} have achieved than their vanilla counterpart on sentence-level and aspect-level sentiment classification. E(commerce)-BERT \cite{ZhangD2020} also demonstrates superior performance on review based QA and review aspect extraction.

\section{Challenges and Future Directions} \label{futuredirection}
In this section, we present the common challenges that face the development of KE-PLMs, and discuss possible research directions to address these challenges.
\subsection{Exploring More Applications}

As discussed in previous section, knowledge enhanced PLMs have achieved success on multiple NLU and NLG tasks. Besides these tasks, there still remain many other applications that can potentially benefit from KE-PLMs such as the following:
\begin{itemize} 
\item \textbf{Text Summarization} KE-PLMs have shown impressive performance on generating more logical and factual text. Factual inconsistency has been an issue that affects the performance of existing document summarization approaches for a long time \cite{KryscinskiW2021}. Adopting the techniques from knowledge enhanced PLMs for improving the factualness of document summarization would be a promising field to explore.
\item \textbf{Machine Translation} How KE-PLMs can improve the quality of machine translation or is not yet explored, to our best knowledge. Integrating linguistic and commonsense knowledge into the PLMs might further their performance.
\item \textbf{Semantic parsing} based applications might also benefit from entity-aware KE-PLMs since it often relies on the entity information in the text.
\end{itemize}

\subsection{Exploring More Knowledge Sources} 

\subsubsection{Application/Domain-specifc Knowledge} Since most existing work focuses on encyclopedia KG, commonsense KG or linguistic knowledge, exploiting domain-specific knowledge is still a relatively under-explored area. As we present in previous sections, several papers have investigated knowledge integration in medical domain and E-commerce domain. As future work, one could investigate the effectiveness of knowledge injection from other domains to PLMs, such as finance, sports and entertainment. It would be exciting to see how KE-PLMs could help improve the NLU and NLG performance in diverse fields.

\subsubsection{Temporal Knowledge Graph} It would be beneficial to study how to effectively integrate temporal knowledge graphs \cite{TrivediR2017} into PLMs. For example, the \textit{personName} in relation triple \textit{(personName, presidentOf, USA)} would not be the same during different time periods. Handling such temporal knowledge could be helpful for further enhancing the capability of KE-PLMs but also poses additional challenges to knowledge representation and integration.

\subsection{Integrating Complex Knowledge More Effectively}

There have been some recent attempts to incorporate knowledge with complex model architecture or perform pretraining in a more sophisticated manner (such as joint training of PLM and KG embedding) \cite{SuY2020}\cite{WangX2019}. However, on benchmark datasets, the reported performance of these more sophisticated schemes does not seem to always outperform approaches that incorporate only entity-level information such as LUKE \cite{YamadaI2020}. Hence, we believe there is still great potential on exploring more effective knowledge integration schemes.

% which are not thoroughly studied in existing literature

\subsection{Optimizing Computational Burden}

Despite the success of KE-PLM on a variety of applications, one should not overlook the fact that incorporating knowledge might incur more computational burden and storage overhead. Most existing work only reports accuracy gains but the incurred cost of knowledge integration has not been thoroughly studied. Designing more time and space efficient solutions would be critical to practical adoption of these knowledge enhanced models.

\subsubsection{Time Overhead}
Many existing methods involves pre-training with auxiliary tasks on large scale linked data \cite{ZhangZ2019}\cite{SunT2020}\cite{YamadaI2020}. How to minimize the amount of pretraining required while still achieving strong performance requires more investigation. Methods such as K-BERT\cite{LiuW2020}, K-Adapter \cite{WangR2020} and Syntax-BERT \cite{BaiJ2021} make progress towards this end by being pretraining-free. We believe there is still potential on further striking a balance between pretraining workload and model performance. When it comes to inference, there is usually additional cost for KE-PLMs as well. For example, a few approaches \cite{JiH2020}\cite{LiuY2020}\cite{SunF2019} involve constructing knowledge subgraphs and learn graph embeddings on the fly, which might increase the inference time. It is worth developing more efficient inference strategy for KE-PLMs to facilitate their adoption in real-world applications.

\subsubsection{Space Overhead}

The additional space consumption of KE-PLMs might also be a concern for their practical deployment. For instance, Fact-as-Experts (FaE) \cite{VergaP2020} builds external entity memory and fact memory (with 1.54 million KB triples) to complement pretrained model. Retrieval Augmented Generation (RAG) \cite{LewisP2020} relies on a non-parametric knowledge corpus with 21 million documents. These integrated entities/facts are not equally useful and some entities might play a more important role in enhancing the model's capability than others. Thus, how to select a succinct set of most important knowledge entries out of a potentially huge space could be a promising future direction.

Another direction worth exploring is to employ model compression techniques \cite{ChengY2018} on both knowledge integration parameters (e.g., entity memory) and language model parameters of KE-PLMs. For example, popular model compression methods such as quantization \cite{ShenS2020}, knowledge distillation \cite{HintonG2015} and parameter sharing \cite{LanZ2020} \cite{DehghaniM2019} can be applied to KE-PLMs for improving time and space efficiency.

% Require building a large external memory for storing knowledge. 

\subsection{Noise Resilient KE-PLMs}

While existing approaches benefit from the additional information provided by knowledge sources, they might also incorporate noise from potentially noisy sources. This is especially true for methods that rely on off-the-shelf toolkit for knowledge extraction:
\begin{itemize}
    \item \textbf{Entity recognition and entity linking}. Existing entity-aware PLMs typically depend on third-party tools or simple heuristics for performing entity extraction and disambiguation. For example, ERNIE \cite{ZhangZ2019}, CoLake \cite{SunT2020}, CokeBERT \cite{SuY2020} use TAGME \cite{FerraginaP2010} to perform entity extraction and linking. ERICA \cite{QinY2020} uses SpaCy to perform NER and then link the entity mentions to Wikidata. \cite{RossetC2020} relies on a (fuzzy) frequency-based dictionary look-up for entity linking. All of these entity linkers would inevitably introduce certain amount of noise to the training process, which might in turn degrade the PLM's performance.
    \item \textbf{Syntax tree} Syntax-BERT\cite{BaiJ2021} and K-Adapter \cite{WangR2020} employ the Stanford Parser\footnote{https://nlp.stanford.edu/software/lex-parser.shtml} for generating dependency parsing trees, which is further used to construct auxiliary loss.
\end{itemize}

%rely on off-the-shelf tools to , Hence, while knowledge is incorporated from these sources, noise is also introduced to the training process:

After the knowledge integration, such noise might propagate to other sub-components of the model and hence negatively impact the performance on downstream tasks. How different noise levels affect the model's performance would be an interesting direction to explore. It would be highly desirable if new KE-PLMs developed in future could demonstrate their robustness to such noise, in both pretraining phase and inference phase.

\subsection{Upper Bound Study}

Recently, gigantic pretrained models with hundreds of billions of parameters, such as Switch Transformers \cite{FedusW2021} and GPT-3 \cite{BrownT2020}, have demonstrated their impressive capability for producing human-like text in a wide range of NLP tasks. Though the prohibitive training and inference cost make them less practical in large-scale real-world applications, they showcase the huge upside of PLMs's performance once equipped with larger training corpus, more model parameters and training iterations.

However, such upper bound study for knowledge enhanced PLMs still seems to be lacking, and there has not been as much effort on pushing the limit for KE-PLMs. It would be interesting to study how well a KE-PLM can perform with massive parameters and knowledge incorporated from a large number of knowledge sources and types (e.g., encyclopedia, commonsense, linguistic). This would be a challenging problem that requires more careful design of knowledge integration schemes, to effectively integrate from multiple knowledge sources and mitigate forgetting. Besides, more advanced distributed training techniques are also critical to empowering the training of such gigantic KE-PLMs.

% \subsection{Lifelong Learning}
%use the information to update the pretrained model and the knowledge graph/fact memory.

%reliability of PTMs is also becoming an issue of great concern with the extensive use of PTMs in production systems. The studies of adversarial attacks against PTMs help us understand their capabilities by fully exposing their vulnerabilities. Adversarial defenses for PTMs are also promising, which improve the robustness of PTMs and make them immune against adversarial a

%\textbf{Reasoning}
%Most existing works focus on incorporating entity information in a relative shallow manner. They are not excellent on reasoning.

\section{Conclusion} \label{conclude}

Integrating knowledge into pretrained language models has been an active research area. We have witnessed an ever-growing interest on this topic, since BERT \cite{DevlinJ2019} popularized the application of PLMs. In this survey, we thoroughly survey and categorize the existing KE-PLMs from a methodological point of view, and establish taxonomy based on knowledge source, knowledge granularity and applications. Finally, we highlight several challenges on this topic and discuss potential research directions in this area. We hope that this survey will facilitate future research and help practitioners to further explore this promising field.

% We also present a wide range of applications that KE-PLMs have demonstrated success. 

%%\bibliographystyle{ACM-Reference-Format}
\bibliographystyle{abbrv}
\bibliography{reference}

\end{document}